\documentclass[journal,twoside,web]{ieeecolor}
\usepackage{generic}
\usepackage{cite}
\usepackage{amsmath,amssymb,amsfonts}
\usepackage{algorithmic}
\usepackage{graphicx}
\usepackage{algorithm,algorithmic}
\usepackage{hyperref}
\usepackage{multirow}
\usepackage{threeparttable}
\usepackage{subcaption}
\usepackage{bm}
\usepackage{amsmath}
\DeclareMathAlphabet{\mathbb}{U}{bbold}{m}{n}

\hypersetup{hidelinks=true}
\usepackage{textcomp}
\def\BibTeX{{\rm B\kern-.05em{\sc i\kern-.025em b}\kern-.08em
    T\kern-.1667em\lower.7ex\hbox{E}\kern-.125emX}}
\begin{document}
\title{Multiple Instance Learning for Glioma Diagnosis using Hematoxylin and Eosin Whole Slide Images: An Indian Cohort Study}
\author{Ekansh Chauhan, Amit Sharma, Megha S Uppin, C.V. Jawahar and  P.K. Vinod
\thanks{We acknowledge IHub-Data, IIIT Hyderabad (H1-002) for financial assistance.}
 % .We acknowledge financial assistance from IHub-Data, a Technology Innovation Hub (TIH) funded by the Department of Science \& Technology (DST), Government of India (GoI) under the National Mission on Interdisciplinary Cyber-Physical Systems (NM-ICPS), International Institute of Information Technology Hyderabad (H1-002)
\thanks{Ekansh Chauhan, Amit Sharma and C.V. Jawahar are with Centre for Visual Information Technology, International Institute of Information Technology Hyderabad, Hyderabad 500032 India  (e-mail: ekansh.chauhan@research.iiit.ac.in, amit.sharma@research.iiit.ac.in, jawahar@iiit.ac.in).}
\thanks{Megha S Uppin is with the Department of Pathology, Nizam's Institute Of Medical Sciences, Hyderabad 500082 India (e-mail: megha\textunderscore harke@yahoo.co.in).}
\thanks{P.K. Vinod is with Center for Computational Natural Sciences and Bioinformatics, International Institute of Information Technology Hyderabad, Hyderabad 500032 India  (e-mail: vinod.pk@iiit.ac.in).}}

\maketitle

\begin{abstract}
 % Brain tumors represent a severe and life-threatening condition, demanding precise diagnosis and tailored treatment strategies. To improve clinical outcomes and enhance patient care, this study presents novel research findings derived from extensive experimentation with various prominent methodologies and combinations of feature extractors and aggregators for key downstream tasks in brain tumor histopathology research. By experimenting with various combinations of these components, we have achieved results that set state-of-the-art benchmarks in multiple downstream tasks for multiple datasets. Furthermore, the study introduces a brain tumor histopathology dataset focusing on the Indian demographic. The best-performing model yields an 88.08 ± 3.98 AUC on 3-way classification of subtypes on our own dataset. It also surpasses the current state-of-the-art reported results on the TCGA dataset by achieving an AUC of 95.81 ± 1.78 and an accuracy of 86.94 ± 3.90 on the subtype classification task. We have also identified a correlation between the assessment methodologies of pathologists and the analogous processes employed by the best-performing model.
 
The effective management of brain tumors relies on precise typing, subtyping, and grading. This study advances patient care with findings from rigorous multiple-instance-learning experimentations across various feature extractors and aggregators in brain tumor histopathology. It establishes new performance benchmarks in glioma subtype classification across multiple datasets, including a novel dataset focused on the Indian demographic (IPD-Brain), providing a valuable resource for existing research. Using a ResNet-50, pretrained on histopathology datasets for feature extraction, combined with the Double-Tier Feature Distillation (DTFD) feature aggregator, our approach achieves state-of-the-art AUCs of 88.08 $\pm$ 3.98 on IPD-Brain and 95.81 $\pm$ 1.78 on the TCGA-Brain dataset, respectively, for three-way glioma subtype classification. Moreover, it establishes new benchmarks in grading and detecting IHC molecular biomarkers (IDH1R132H, TP53, ATRX, Ki-67) through H\&E stained whole slide images for the IPD-Brain dataset. The work also highlights a significant correlation between the model’s decision-making processes and the diagnostic reasoning of pathologists, underscoring its capability to mimic professional diagnostic procedures.

 \end{abstract}


\section{Introduction}
\label{sec:introduction}

\IEEEPARstart{B}{rain} tumors, known for their lethality, represent a critical medical condition characterized by the abnormal growth of cells within the central nervous system (CNS), which can lead to severe neurological dysfunction. The World Health Organization (WHO) Classification of CNS tumors (WHO CNS 5) includes various types and subtypes based on their histologic and molecular genetic characteristics \cite{louis20212021}. Due to genetic predisposition, environmental factors, and lifestyle-related variables in India, the occurrence of CNS tumor cases fluctuates between 5 and 10 cases per 100,000 individuals \cite{dasgupta2016indian, nairindiasummary}. Glial tumors are the most common type of brain tumor, primarily Astrocytoma, Glioblastoma, and Oligodendroglioma \cite{Thakur_Gahine_Kulkarni_2018, hanif2017glioblastoma, tamimi2017epidemiology}. Glioblastoma, the most aggressive among them, which accounts for approximately 57\% of all gliomas, represents a highly malignant glioma, with a median survival of just 12-15 months, even with aggressive treatment approaches and associated with a high rate of recurrence due to its infiltrative nature \cite{stupp2005radiotherapy,ostrom2021cbtrus}. Astrocytoma, on the other hand, originates from astrocytes and is graded based on its degree of malignancy, low or high. Oligodendroglioma, a rarer subtype, arises from oligodendrocytes and is often characterized by genetic alterations, which can influence treatment strategies and predict better outcomes \cite{louis20162016}.

Histopathology is a fundamental branch of medical science involving microscopic examination of tissue specimens to diagnose diseases, specifically cancers, and understand the underlying cellular and structural abnormalities. Traditionally, it involves pathologists visually inspecting stained tissue sections under a microscope. Hematoxylin and Eosin (H\&E) staining and Immunohistochemistry (IHC) staining are two standard techniques in Histopathology that serve distinct purposes in medical research and diagnostics. Hematoxylin stains cell nuclei blue-purple, whereas Eosin stains cytoplasm and extracellular matrix various shades of pink. H\&E staining is essential for assessing tissue structure, identifying cellular components, and diagnosing subtypes and corresponding WHO grade \cite{louis20212021}. 
Glioma grading is essential for determining patient prognosis and guiding treatment plans. The differentiation between high-grade gliomas (HGG) and low-grade gliomas (LGG) is necessary, as it indicates variations in tumor aggressiveness and correlates with patient survival rates \cite{Cho2018, jin2021artificial, KER2019239, BARKER201660}. Thus, along with three-way subtype classification, we also focus on distinguishing between HGG and LGG to assess tumor progression and inform therapy planning accurately. 

% Hence, the division of grades between high-grade gliomas (HGG) and low-grade gliomas (LGG), which has been widely adopted in the community \cite{Cho2018, jin2021artificial, KER2019239, BARKER201660}, is necessary. Thus, along with 3-way subtype classification, we also focuses on differentiating HGG from LGG for assessing tumor progression and therapy planning. 

%%%%%%%%%%%%%%%%%%%%%%%%%%%%%%%%%%%%%%%%%%%%%%%%%%%%%%%%%%%%%%%%%%%%%%%%%%%%%%%%%%%%%%%%%%%%%%%%%%%%%%%%%%%%%%%%%%%%%%%%%%%%%%%%%%%%%%%%%%%%%%%%%%%%%%%%%%%%%%%%%%%%%%%%%%%%%%%%%%%%%%%%%%%%%%%%%%%%%%%%%%%%%%%%%%%%%%%%%%%%%%%%%%%%%%%%%%%%%%%%%%%%%%%%%%%%%%%%%%%%%%

\begin{figure*}[t]
    \centering
    \includegraphics[width=\textwidth]{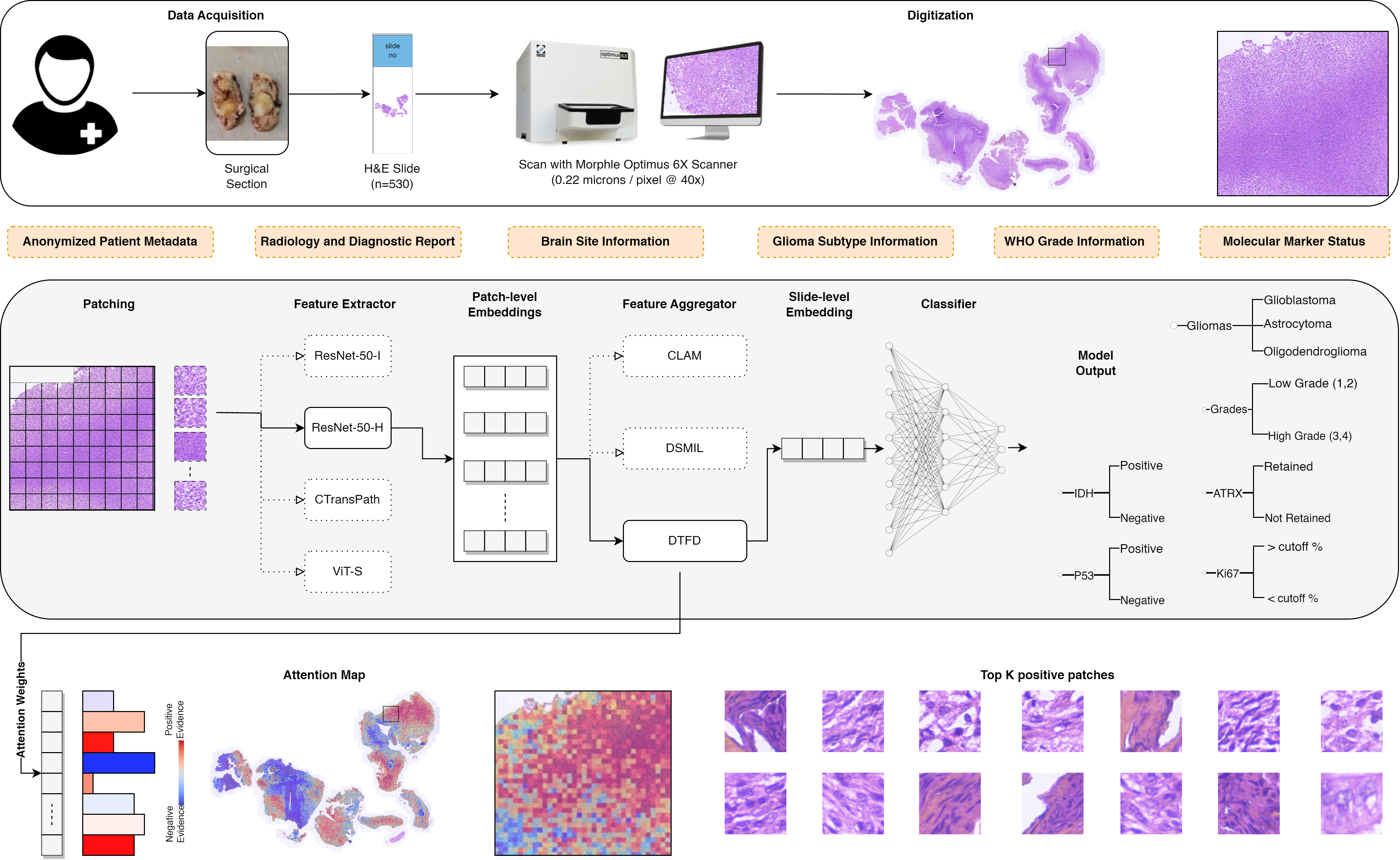}
    \caption{Multi-Instance Learning Framework for Subtype Classification in Brain Histopathology. Tissue from surgical samples is digitized and processed into patches, and patch-level features are extracted using a pre-trained feature extractor network. Subsequently, the feature aggregation method pools the patch-level representation into slide-level representation, which is then classified to determine tumor type, grade, and molecular markers. Then, attention mapping is used for interpretability, providing critical insights into the neural network's decision-making process.}
    \label{fig:ipd_pipeline}
\end{figure*}

Additionally, IHC staining is a specialized technique to detect specific proteins or antigens within tissue samples. It involves using antibodies tagged with chromogenic or fluorescent markers to visualize the presence and distribution of specific proteins in tissues. IHC is valuable for investigating biomarkers like isocitrate dehydrogenase (IDH / IDH1R132H), alpha thalassemia/mental retardation syndrome X-linked (ATRX), TP53, Ki-67 and identifying specific cellular components related to diseases. In this study, we explored the possibility of predicting the status of IHC biomarkers (IDH, ATRX, TP53, and Ki-67) using H\&E stained whole slide images (WSIs). This will help to understand between morphology and genetics of the tumor and perhaps help in assisting the diagnosis and reducing multiple molecular assessments using IHC. 

% To mimic the diagnosis of pathologists, we treated Ki-67, the percentage of positively stained cells among the total number of cancer cells assessed as a binary classification task by varying thresholds.

In recent years, the field of histopathology has undergone a significant transformation, primarily due to the emergence of digital histopathology \cite{irshad2013methods}, the available datasets like The Cancer Genome Atlas (TCGA) \cite{tcga_data}, and the application of deep learning methods for their analysis \cite{lu2021data, tellez2019neural}. This evolution in brain tumor research domain underscores the importance of diverse datasets that extend beyond the confines of American or European populations. The global landscape of brain tumors demands a comprehensive understanding facilitated by including datasets from various demographics. Deep learning methods have demonstrated remarkable efficiency in processing extensive histopathological data with unprecedented speed and accuracy.

While existing research is able to predict the subtype, grading, and biomarkers directly from routine H\&E stained histopathology slides for many tumor types \cite{Campanella2019, ilse18a, lu2021data}, including brain tumor \cite{Liu2020, brain_subtype_grade}, showing the potential to accelerate diagnostic workflows and reduce the costs associated with molecular testing. However, a limited focus is still given to brain tumor classification based on the updated WHO guidelines (the fifth edition, WHO CNS5, 2021) \cite{Who_2021}. Due to the large size of histopathology images and lack of patch-level labels in a WSI, the Multiple Instance Learning (MIL) framework has been widely adopted in the field of digital pathology. MIL is a two-stage framework: (i) Feature extractor and (ii) Feature Aggregator. The use of standard ImageNet pre-trained weights and large-scale domain-aligned (pathology) pre-training via self-supervised learning (SSL) \cite{Kang_2023_CVPR, WANG2022102559, Chen_2022_CVPR} for extracting latent features of patches underscores its effectiveness on various downstream pathology tasks. To aggregate or pool these features, Campanella et al. \cite{Campanella2019} employed an RNN-based approach for feature-level aggregation in the classification of clinical pathology whole slide images. ABMIL, as introduced by Ilse et al. \cite{ilse18a}, leverages attention mechanisms to determine instance weights through a trainable neural network. This method enhances traditional MIL pooling operations, such as max or mean pooling, by aggregating bag-level representations based on weighted averages of instance-level features. Expanding on the concept of attention-based instance scoring, Clustering-constrained Attention MIL (CLAM), developed by Lu et al. \cite{lu2021data}, delves into multi-class predictions using instance-level clustering, thereby facilitating the learning of class-specific features. Dual-stream MIL (DS-MIL) \cite{Li_2021_CVPR} employs non-local attention for aggregation to learn classifiers. A Transformer-based Multiple Instance Learning approach (TransMIL) \cite{shao2021transmil} explores the relationships between instances in terms of their morphology and spatial characteristics using the Nyström method for approximating all-to-all self-attention, though it requires significant computation resources. Double-Tier Feature Distillation (DTFD) MIL, introduced by Zhang et al. \cite{Zhang_2022_CVPR}, introduces the concept of "pseudo-bags" addressing issues associated with small cohorts. Another contribution is a transformer-based hierarchical ViT proposed by \cite{Chen_2022_CVPR}, focusing on using spatial relationships among instances. However, despite these advancements, the downstream classification tasks involving brain tumor subtypes and the identification of IHC biomarkers remain underexplored.

Our research addresses a critical gap in brain tumor studies by introducing a glioma subtype dataset within the India Pathology Dataset (IPD) consortium.\footnote{https://hai.iiit.ac.in/ipd/} This dataset facilitates research in India and encourages global collaboration to advance our understanding and combat this severe illness. Fig.~\ref{fig:ipd_pipeline} shows the pipeline followed in this work. Through rigorous experimentation, we aim to identify a specific combination of feature extractor and aggregator that performs well for various tasks based on the WHO CNS 5 guidelines. The outcomes of our study emphasize the opportunity to advance the state of the art in deep learning applications for brain tumor classification.

The main contributions of our work are as follows:

\begin{itemize}
    \item  We created a dataset tailored to the Indian region and one of the largest collections of brain WSIs globally, providing a valuable resource for research in this domain.

    \item To our knowledge, this is the first study predicting IHC molecular biomarkers using cost-effective H\&E slides, eliminating the need for expensive IHC staining. This enhances accessibility and also explores the potential of H\&E stains in accurately determining these biomarkers.
    
    \item We achieve state-of-the-art (SOTA) results on TCGA data for glioma subtype classification by using a specific self-supervised pre-trained feature extractor and a feature aggregator.

    \item We benchmark the IPD-Brain Dataset for six subtype classification tasks, i.e., glioma subtype classification, grading, IDH, ATRX, TP53, and Ki-67.

    \item We introduce Ki-67, the percentage of positively stained cells among the total number of cancer cells assessed, as a binary classification task by varying thresholds.

    \item Through rigorous experimentation with all combinations of four feature extractors and three aggregators, we provide insights into the most effective approaches for analyzing brain histology data.
    
\end{itemize}

% The remainder of this work is organized as follows:
% Section II describes our IPD-Brain dataset description and associated statistics. Section III formulates the problem mathematically and discusses the existing methods used as baselines and benchmarking. Section IV presents the experiments, and Section V shares the results. Section VI explores the explainability of the best-performing model qualitatively. Section VII provides the conclusions and future work.

\section{Dataset Description}

\begin{figure}[h]
    \centering
    \begin{subfigure}{0.24\textwidth}
        \centering
        \includegraphics[width=\linewidth]{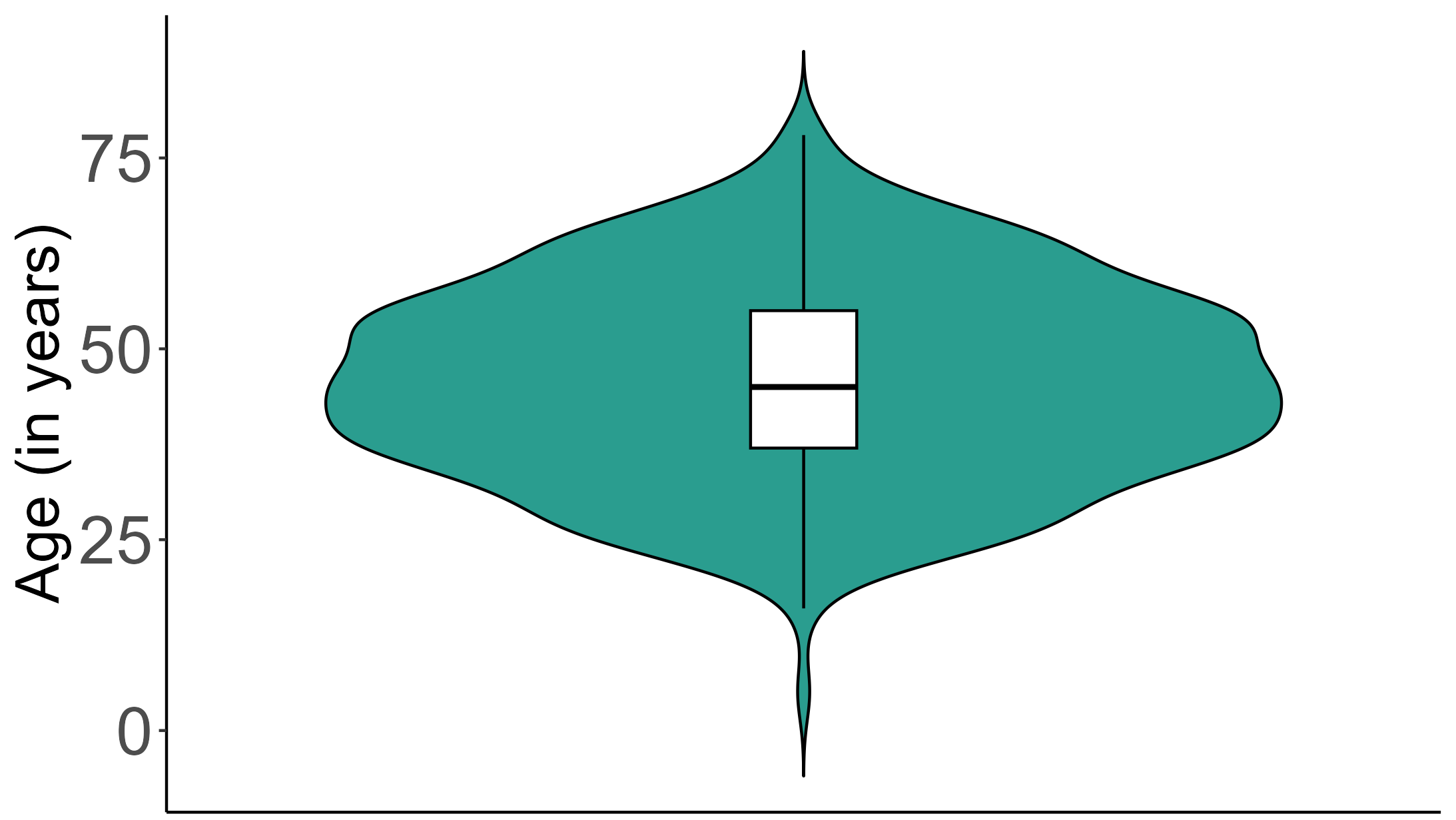}
        \caption{Age}
    \end{subfigure}
    \hfill
    \begin{subfigure}{0.24\textwidth}
        \centering
        \includegraphics[width=\linewidth]{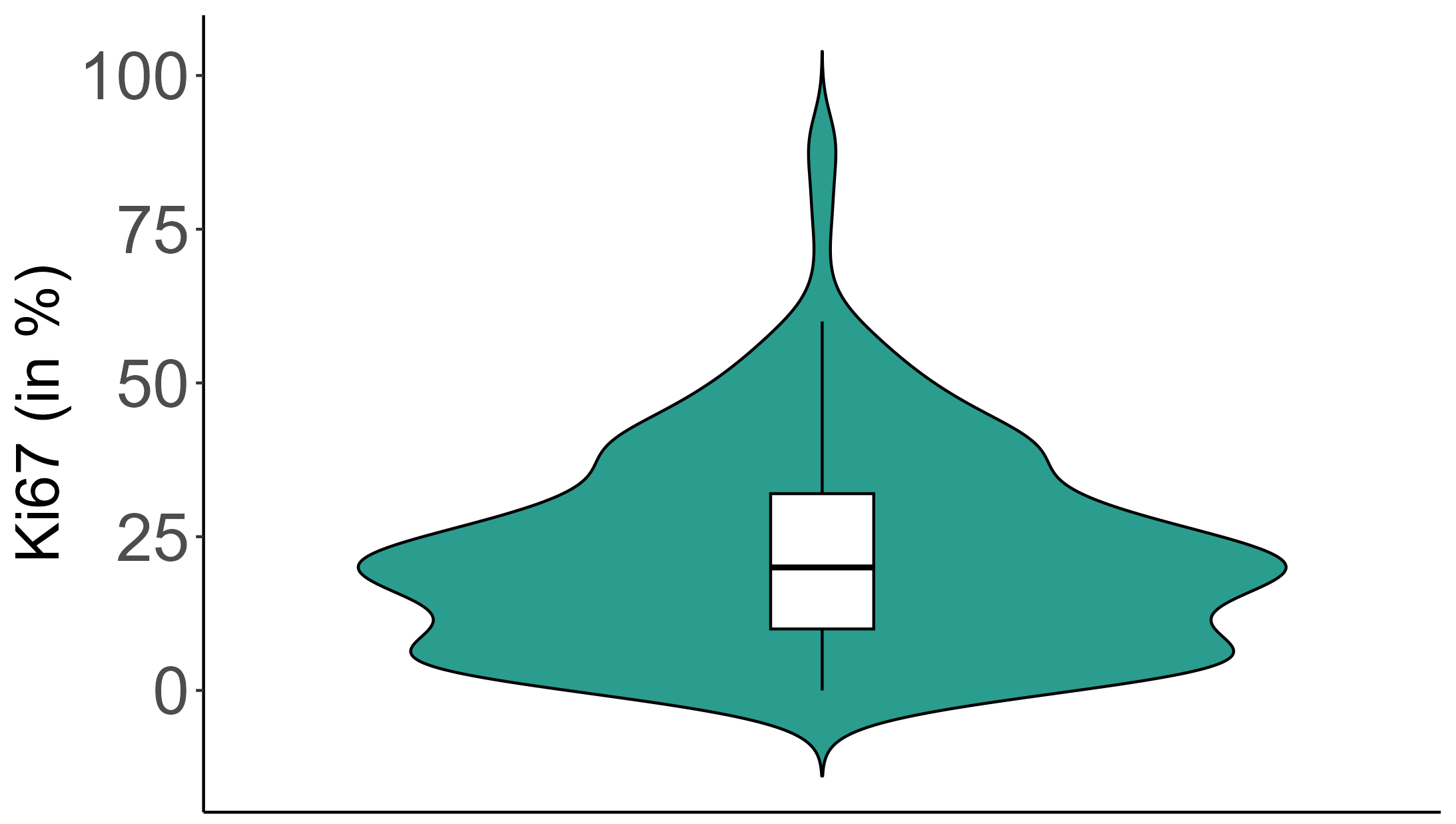}
        \caption{Ki-67}
    \end{subfigure}
    
    \begin{subfigure}{0.24\textwidth}
        \centering
        \includegraphics[width=\linewidth]{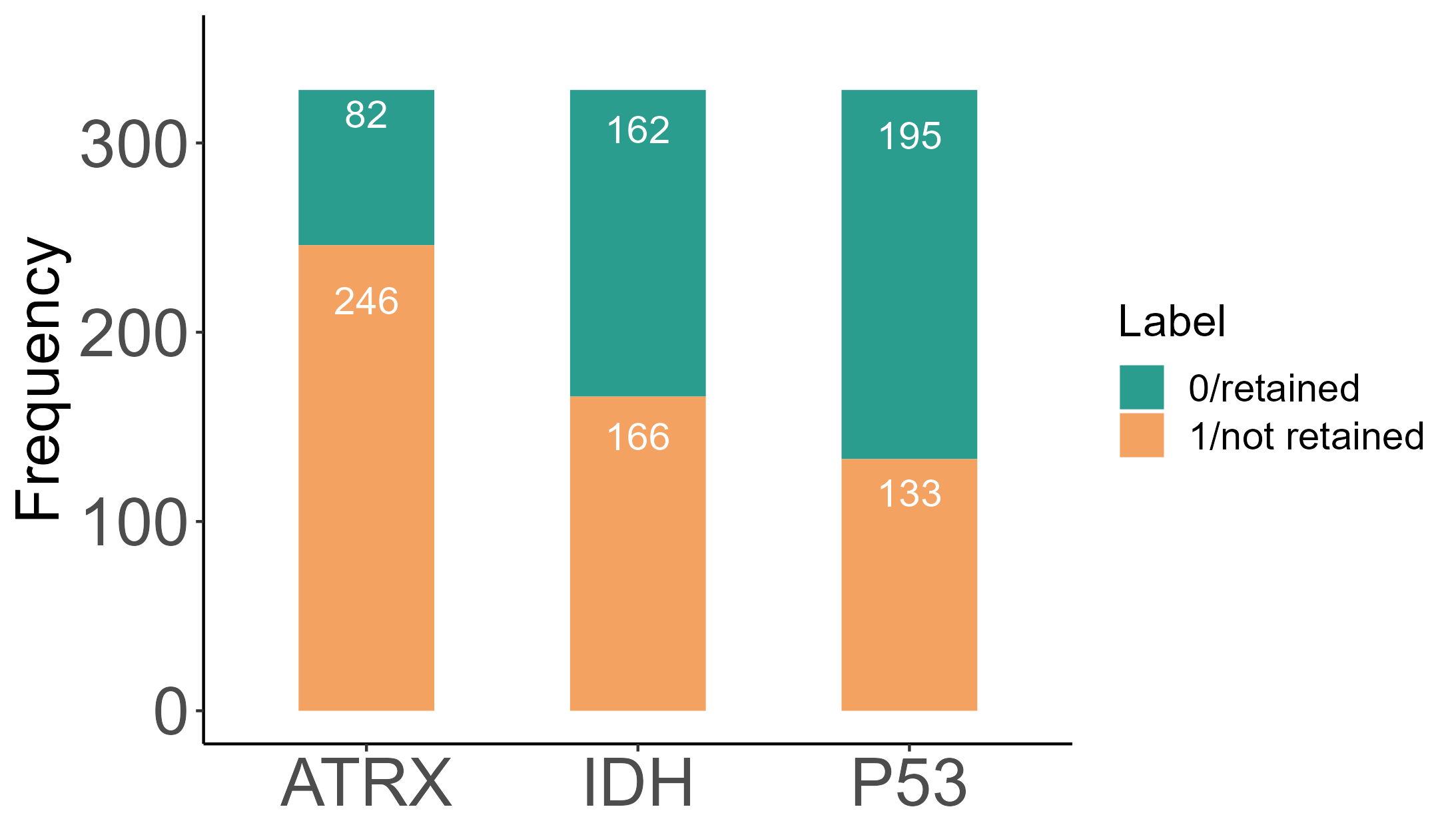}
        \caption{IHC biomarkers}
    \end{subfigure}
    \hfill
    \begin{subfigure}{0.24\textwidth}
        \centering
        \includegraphics[width=\linewidth]{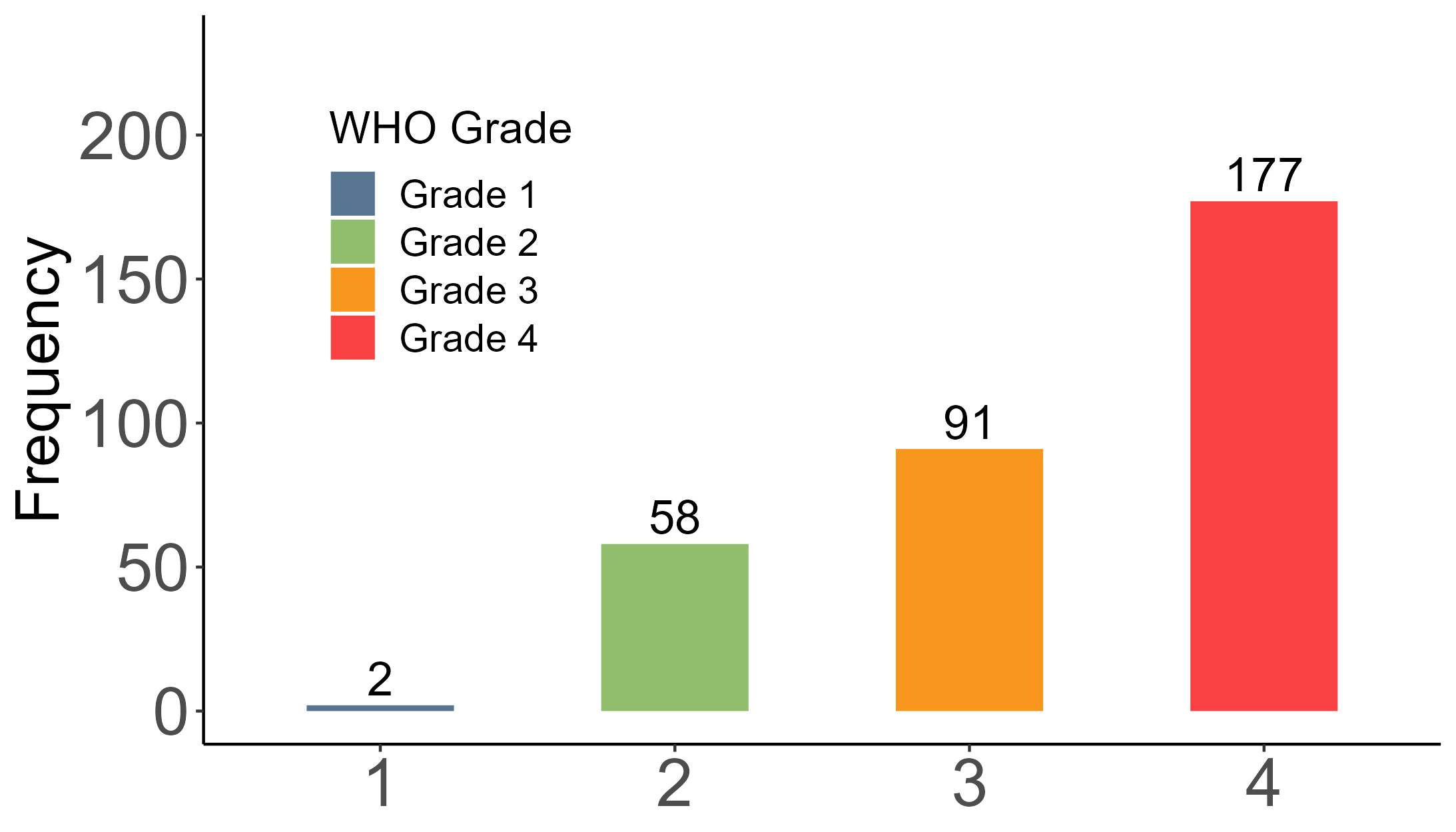}
        \caption{Grade}
    \end{subfigure}
    \caption{Distribution of phenotypic information and labels in IPD-Brain.}
    \label{fig:dataset}
\end{figure}

\begin{figure}[h]
    \centering
    \includegraphics[width=0.45\textwidth]{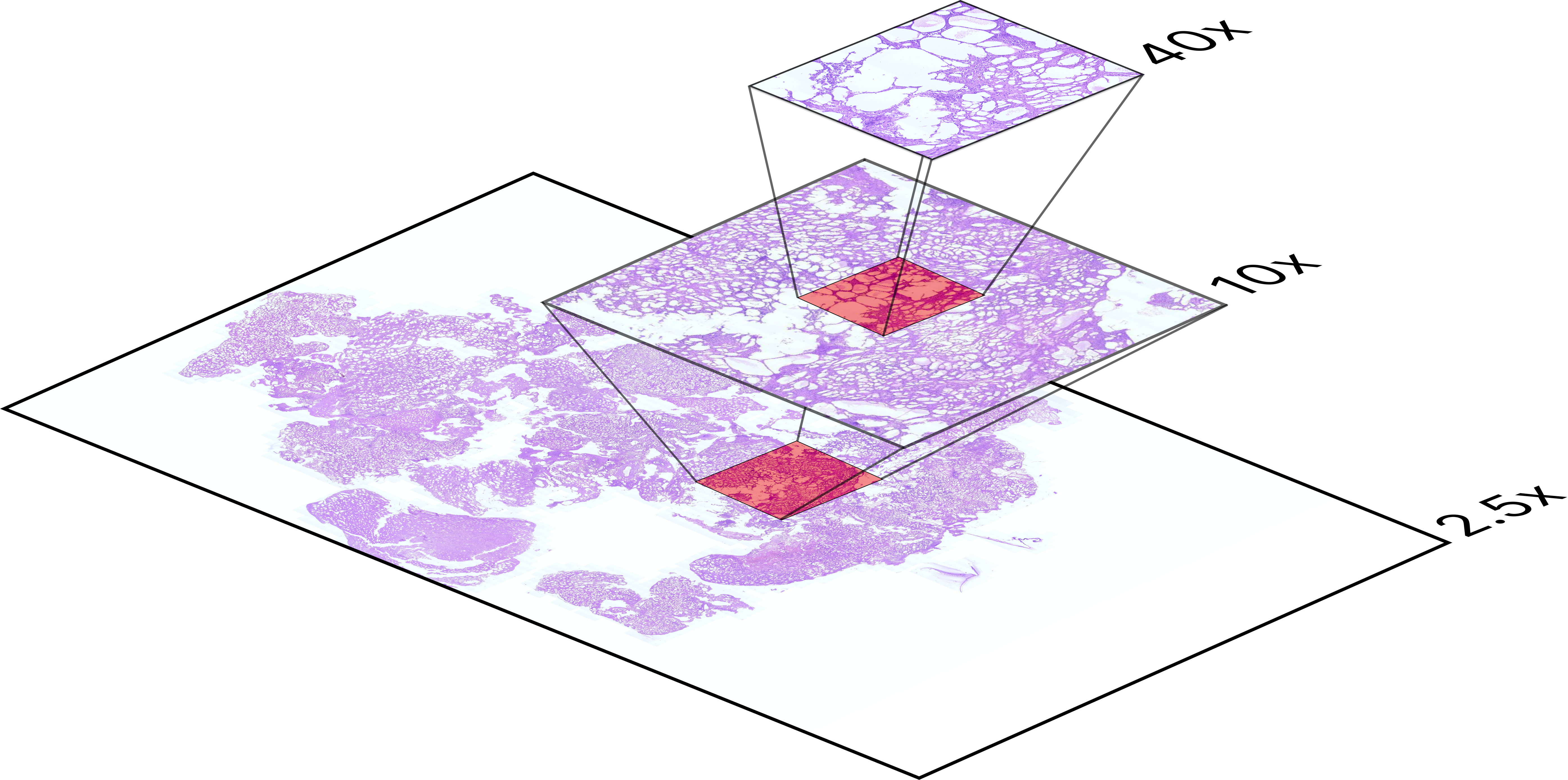}
    \caption{IPD-Brain Dataset sample at various magnifications demonstrate the enhanced detail from staining and digitization, crucial for deep learning analysis.}
    \label{fig:dataset_snapshots}
\end{figure}

As a part of the India Pathology Dataset (IPD) consortium, in this study, 530 digitized slides from 367 patients (retrospective and prospective cases) were acquired at the Department of Pathology of Nizam’s Institute Of Medical Sciences (NIMS), Hyderabad, from a diverse patient cohort presenting three brain tumor types (Glioblastoma, Astrocytoma, Oligodendroglioma). Table I and Fig.~\ref{fig:dataset} summarize multiple statistics of the IPD-Brain (in-house) dataset.  Each specimen was collected following stringent ethical protocols and quality control measures to ensure representativeness and integrity. The collection process was closely monitored to maintain the tissue’s physiological state, minimizing any alterations that might affect the histopathological outcomes. A sample of this dataset at multiple magnification levels can be seen in Fig.~\ref{fig:dataset_snapshots}. 

\begin{table}[t]
\centering
\caption{Statistics of IPD-Brain histopathology dataset}
\label{tab:ipd-brain}
\begin{tabular}{|lll|}
\hline
\multicolumn{1}{|l|}{} & \multicolumn{1}{l|}{Patient level} & Slide level\\ \hline
\multicolumn{3}{|c|}{Scan Statistics:} \\ \hline
\multicolumn{1}{|l|}{Total Scanned WSIs} & \multicolumn{1}{l|}{367} & 530 \\ \hline
\multicolumn{1}{|l|}{\begin{tabular}[c]{@{}l@{}}Scanned WSIs w/complete EHR record\end{tabular}} & \multicolumn{1}{l|}{328} & 484 \\ \hline
\multicolumn{3}{|c|}{Glioma Subtype Statistics:} \\ \hline
\multicolumn{1}{|l|}{Glioblastoma (GBM)} & \multicolumn{1}{l|}{168 (51.22\%)} & 246 (50.83\%) \\ \hline
\multicolumn{1}{|l|}{Astrocytoma (A)} & \multicolumn{1}{l|}{88 (26.83\%)} & 133 (27.48\%) \\ \hline
\multicolumn{1}{|l|}{Oligodendroglioma (O)} & \multicolumn{1}{l|}{72 (21.95\%)} & 105 (21.69\%) \\ \hline
\multicolumn{3}{|c|}{Phenotypic Statistics of Glioma Patients:} \\ \hline
\multicolumn{1}{|c|}{Sex:} & \multicolumn{2}{l|}{} \\ \hline
\multicolumn{1}{|l|}{Female} & \multicolumn{2}{l|}{136 (41.46\%)} \\ \hline
\multicolumn{1}{|l|}{Male} & \multicolumn{2}{l|}{192 (58.54\%)} \\ \hline
\multicolumn{1}{|c|}{Age} & \multicolumn{2}{c|}{45.89 $\pm$ 12.93} \\ \hline
\multicolumn{1}{|l|}{Female} & \multicolumn{2}{l|}{45.37 $\pm$ 12.64} \\ \hline
\multicolumn{1}{|l|}{Male} & \multicolumn{2}{l|}{46.27 $\pm$ 13.15} \\ \hline
\end{tabular}
\end{table}

% The process of preparing tissue samples for histopathological examination involves several meticulous steps. Initially, tissues are fixed in a 10\% neutral buffered formalin solution to preserve cellular structures and stop enzymatic breakdown. The fixation duration is adjusted based on tissue type to avoid overfixation and preserve histological details. Following fixation, tissues are dehydrated with increasing concentrations of alcohol, preparing them for embedding in paraffin wax. This embedding process ensures tissues are well-supported for precise sectioning. Thin sections (4-5 micrometers) are then cut using a microtome, allowing for effective light transmission and stain penetration during microscopic examination. Finally, sections undergo H\&E staining, where Hematoxylin stains cell nuclei blue-purple and Eosin stains cytoplasmic components pink-red, highlighting various tissue components and aiding in the identification of any pathological changes.

The digitization of stained slides into high-quality digital images was accomplished using the Morphle DigiPath 6T Scanner, capturing details at a 40x magnification akin to conventional microscopy. The digital images were saved in the TIFF format. Each tiff file was also enriched with metadata detailing scan specifics and magnification, enhancing the data's utility. Furthermore, NIMS provided electronic health records (EHRs) for all slides, identifying tumor subtypes, grades, and IHC biomarkers, complemented by extensive metadata that includes patient demographics, radiology, and diagnostic reports, thereby offering a nuanced understanding of each specimen within its clinical context.

This India region-specific dataset is developed to support global collaboration in brain tumor research, enhancing diversity among existing cohorts and providing insights into potential regional and ethnic variations in the disease.

% This process ensured accurate color and contrast representation of H\&E stains, with each slide being thoroughly scanned to generate whole slide images (WSI) for comprehensive and focused tissue examination. Quality assurance was rigorously maintained to eliminate any potential distortions like focus errors or dust. 

\section{Methodology}

\subsection{MIL Problem formulation}
In histopathology classification, MIL is a compelling solution to the challenge of sparse and imprecise annotations often encountered in medical image analysis. Unlike traditional supervised learning, which relies on precisely labeled instances (patches), MIL operates on 'bags' of instances, with the label determined by the collective characteristics of these instances (slide-level). Here, the mathematical formulation of the MIL framework for subtype classification is extended and defined as follows:

Consider a dataset \(X\) comprising \(M\) WSIs, denoted as \(X = \{X_1, X_2, ..., X_M\}\). Each WSI is a 'bag' in MIL terminology, representing an entire histopathology slide. The label of the \(m\)-th WSI, denoted as \(Y_m\), in an n-way subtype classification task, is determined based on the subtype presence within the slide, with each slide categorized into one of \(n\) possible subtypes or negative slide. The instances within a bag, denoted as \(x_i\), are individual image patches or regions extracted from the histopathology slide. For a given \(m\)-th WSI, the instances are represented below in \eqref{eq:x_m}:

\begin{equation}
    X_m = {x_1, …, x_i, …, x_{p-1}, x_p}
\label{eq:x_m}
\end{equation}

where $p$ is the number of patches in the $m$-th WSI, and each $ x_i \in \mathbb{R}^{256 \times 256 \times 3} $. Although the exact labels $y_i$ for these instances are unknown in MIL, they are implicitly associated with the overall label $Y_m$, \eqref{eq:y_m}, of the bag.

\begin{equation}
Y_m =
\begin{cases}
k, & \text{if } \exists i \in p: y_i = k \\
0, & \forall i \in p: y_i = 0
\end{cases}
\label{eq:y_m}
\end{equation}

Here, $k \in n$ represents the subtype class labels. This implies that the bag belongs to class $k$ if any instance within it is determined to belong to that class.  Otherwise, if all the instances in a bag are negative, the bag is considered negative (i.e., belongs to the negative class - $0$). In this work, we only focused on positive subtype classes.

The feature extractor function, denoted as $f$ (section:~\ref{f}), is crucial in transforming each instance $x_i$ into a deep meaningful representation. It maps each instance to a fixed-dimensional feature vector $d$, expressed as \eqref{eq:f_xi}.

\begin{equation}
    f(x_i) = v_i
\label{eq:f_xi}
\end{equation}

where $v_i \in \mathbb{R}^{d \times 1}$ is the feature vector of size \(d\) extracted from the \(i\)-th instance \(x_i\). The function \(f\) could be implemented through various methods, including handcrafted feature extractors, deep CNNs, or ViTs.

A bag $B \in \mathbb{R}^{p \times d}$ represents the entire histopathology slide and is defined as a collection of instance feature vectors. Formally, the bag is constructed through the concatenation of the feature vectors of its instances, expressed using the concatenation symbol \(\oplus\) in \eqref{eq:b}.

\begin{equation}
    B = \bigoplus_{i=1}^p v_i
\label{eq:b}
\end{equation}

The goal is to learn a hypothesis function $g$ (section:~\ref{g}) that effectively maps a bag to its corresponding label. This function can be viewed as a combination of an Aggregator and a Classifier. The aggregator transforms the representation of the bag by using any aggregation functions, such as the average or maximum of instance features or more complex attention-based aggregation methods. The output of the Classifier is the predicted class out of $n$ classes.

\subsection{WSI Preprocessing}
The process begins by segmenting tissue regions within digitized slides. The image is converted to the HSV color space, and a mask is created by thresholding the saturation channel after smoothing. Morphological operations fill gaps and holes, and contours are filtered based on area. We have included optimal segmentation parameters for our dataset in section~\ref{experiment}. Following segmentation, the algorithm crops 256x256 patches from within the segmented regions based on user-specified magnification.

\subsection{Feature Extractors (f)}
\label{f}
While ImageNet pre-trained weights are commonly utilized in histopathology tasks, we expanded our comparative analysis to encompass various SSL methods recognized for their strong performance. The selection of multiple SSL approaches was motivated by the absence of a definitive superior way across all settings; apart from the baseline ResNet-50 pre-trained on ImageNet, our evaluation considered high-performance networks with domain-specific pre-training (Histopathology). These networks included a ResNet-50 model pre-trained via the Barlow Twins method, a Vision Transformer model pre-trained using the DINO method, and CTransPath, a hybrid model merging a convolutional neural network (CNN) with a multiscale Swin Transformer architecture, pre-trained through semantically-relevant contrastive learning. By incorporating these diverse options, we aimed to comprehensively evaluate their generalizability to our in-house histopathology dataset.

\subsection{MIL Aggregation Models (g):}
\label{g}
We also employed three feature aggregator models selected based on their performance across various studies \cite{lu2021data, Li_2021_CVPR, Zhang_2022_CVPR}. These aggregation models leverage attention-based learning to automatically identify the subtype or diagnostically significant sub-regions for precise whole-slide classification.

\subsubsection{CLAM}
It uses gated attention to assign scores to patches within a whole slide, leveraging these scores for slide-level aggregation through a weighted average for multi-class classification. It operates with n parallel attention branches for n class-specific representations, though studies show the single-branch configuration often performs better. It incorporates domain knowledge for additional guidance in instance-level clustering.

\subsubsection{DSMIL}
It employs a dual-stream architecture with self-supervised contrastive learning, for instance embeddings. One stream identifies the key instance, whereas the other aggregates embeddings for a comprehensive bag representation. It uses a unique approach to weight instance contributions based on similarity to the key instance, enhancing the final bag score through a sophisticated attention mechanism.

\subsubsection{DTFD}
It generates pseudo-bags from bag divisions, maintaining the original label for each. It utilizes a two-tier attention-based MIL model, first estimating pseudo-bag probabilities, then deriving instance probabilities and distilled features for final parent bag likelihood prediction. Despite potential noise, it leverages deep neural network resilience, focusing on Aggregated Feature Selection (AFS) for feature distillation in this context.

\subsection{Transfer Learning for IHC Biomarkers}
Furthermore,  our investigation aimed to assess "Is H\&E stained slide enough?" by determining the capability of our weakly-supervised deep learning subtype classification model in predicting common diagnostic molecular markers from glioma H\&E slides. We sought to evaluate its potential utility as a clinical secondary check or a cost-effective alternative to actual molecular biology experiments. Using transfer learning, a technique that adapts a model trained on one task to a related new task, we systematically explored the effectiveness of the best-performing model, initially trained for tumor subtype classification, in inferring molecular biomarkers within a weakly supervised context. Fine-tuning was conducted on the three-class subtyping model for each of the three binary classifications of IHC mutations (somatic mutation versus wild type for IDH1, TP53, and ATRX mutations) and one classification task at various cutoffs for assessing tumor cell proliferation (Ki-67).

\section{Experimentation}
\label{experiment}

\subsection{Experiment Setup}

In our study, we conducted comprehensive experiments using two datasets: the in-house IPD-Brain Dataset and the publically available TCGA-Brain. This section delineates the experimental setup employed. For feature extraction at the patch level from WSIs, we used four distinct networks: ResNet-50 pre-trained on ImageNet in a supervised manner, ResNet-50 pre-trained using the Barlow Twins SSL method for histopathology, CTransPath pre-trained with semantically relevant contrastive learning for histopathology, and ViT-S pre-trained via the DINO approach for histopathology.

Upon extracting features using these models, we applied three aggregation methods - CLAM, DSMIL, and DTFD - for the downstream classification task. We identified the most effective extractor-aggregator combination on the IPD-Brain dataset for the three-way subtype classification task and subsequently extended its application for further experimental analysis. This involved first testing the chosen combination on the TCGA-Brain dataset for subtype classification, followed by employing the same combination for classifying grading (LGG vs HGG) and IHC biomarkers mutations (TP53, IDH, ATRX, Ki-67) using H\&E stained slides. This systematic approach allowed us to assess the efficacy of our models rigorously across different datasets and tasks for brain histopathology.

% implementing all cris cross, setting up baselines. Ten fold. 80-10-10 split information, Two datasets, ensuring patient-wise division, training is terminated after n epochs of no decrease in validation loss. Official implementations of the methods were used when available, and every effort was made to reproduce the results—magnification level. 

\subsection{Evaluation Metrics}

The primary performance metric used for all the experiments is the Area Under Curve (AUC), which offers a comprehensive assessment and is less susceptible to class imbalance. Additionally, we consider slide-level accuracy (Acc) and the macro-averaged F1 score. These metrics collectively provide a holistic view of model performance: the AUC score indicates the ability to differentiate between classes, whereas the F1 score offers insight into the actual performance in the context of imbalanced data.

\subsection{Implementation Details}

The preprocessing of WSIs was done at 40x magnification. We used CLAM preprocessing steps with median blurring using a 9x9 kernel, followed by Otsu thresholding. Morphological closing with a 5x5 kernel was applied for finer segmentation. Tissue contours were identified with an area filter threshold of 80, while holes within these contours were managed with a threshold of 10 and a limit of 20 holes per contour.

We adhered to official implementations of the methods wherever possible, exerting diligent efforts to replicate the code accurately. A comprehensive evaluation was conducted for each combination of feature extractors and aggregators. The dataset was divided into training, validation, and testing phases with proportions of 80\%, 10\%, and 10\%, respectively, on a patient-wise basis. This approach ensured data leakage prevention, meaning all slides from a single patient were exclusively allocated to one of these phases. Additionally, we implemented a 10-fold cross-validation for each experimental setup.

\begin{table*}[htbp]
\centering
\caption{Tumor subtype classification on IPD-Brain}
\label{tab:subtype_ipd}
\begin{tabular}{|l|l|l|l|l|l|l|}
\hline
% Feature Extractor & Dataset Category & SSL Method & Feature Aggregator & AUC & ACC & F1 \\ \hline
% F & D & SSL & G & AUC & ACC & F1 \\ \hline
\begin{tabular}[c]{@{}l@{}}Feature \\ Extractor\end{tabular} & \begin{tabular}[c]{@{}l@{}}Dataset \\ Category\end{tabular} & SSL Method & \begin{tabular}[c]{@{}l@{}}Feature \\ Aggregator\end{tabular} & AUC & ACC & F1 \\ \hline

\multirow{3}{*}{Resnet-50} & \multirow{3}{*}{Imagenet} & \multirow{3}{*}{Supervised} & CLAM-sb & 78.85 $\pm$ 6.27 & 65.9 $\pm$ 6.72 & 56.86 $\pm$ 12.79\\ \cline{4-7} 
 &  &  & DSMIL & 79.72 $\pm$ 5.78 & 69.29 $\pm$ 6 & 61.37 $\pm$ 7.47 \\ \cline{4-7} 
 &  &  & DTFD & 80.5 $\pm$ 3.88 & 75.62 $\pm$ 4.11 & 69.78 $\pm$ 5.08\\ \hline
\multirow{3}{*}{Resnet-50} & \multirow{3}{*}{Histopathology} & \multirow{3}{*}{Barlow Twin} & CLAM-sb & 84.55 $\pm$ 4.09 & 73.27 $\pm$ 5.38 & 67.61 $\pm$ 7.03\\ \cline{4-7} 
 &  &  & DSMIL & 80.87 $\pm$ 2.56 & 67.12 $\pm$ 5.95 & 59.98 $\pm$ 6.80\\ \cline{4-7} 
 &  &  & \textbf{DTFD} & \textbf{88.08 $\pm$ 3.98} & \textbf{79.83 $\pm$ 3.31} & \textbf{75.87$\pm$ 4.79} \\ \hline
\multirow{3}{*}{CTransPath} & \multirow{3}{*}{Histopathology} & \multirow{3}{*}{SRCL} & CLAM-sb & 83.33 $\pm$ 3.63 & 72.08 $\pm$ 5.70 & 68.11 $\pm$ 5.57\\ \cline{4-7} 
 &  &  & DSMIL & 83.32 $\pm$ 5.16 & 69.73 $\pm$ 7.11 & 65.08 $\pm$ 7.68 \\ \cline{4-7} 
 &  &  & DTFD & 87.12 $\pm$ 3.96 & 78.75 $\pm$ 5.78 & 75.73 $\pm$ 6.43 \\ \hline
\multirow{3}{*}{ViT-S} & \multirow{3}{*}{Histopathology} & \multirow{3}{*}{DINO} & CLAM-sb & 82.95 $\pm$ 5.04 & 69.88 $\pm$ 6.96 & 65.2 $\pm$ 7.17 \\ \cline{4-7} 
 &  &  & DSMIL & 81.96 $\pm$ 6.06 & 68.84 $\pm$ 5.56 & 62.61 $\pm$ 7.64 \\ \cline{4-7} 
 &  &  & DTFD & 87.67 $\pm$ 4.14 & 78.89 $\pm$ 3.98 & 75.41 $\pm$ 3.50 \\ \hline
\end{tabular}
\end{table*}

Training was conducted for a minimum of 50 epochs and a maximum of 200 epochs, with early stopping implemented with a patience of 25 epochs on validation loss. The Adam optimizer was used with a batch size of 1. Specific learning rates and weight decay parameters were set for each aggregator: DSMIL at 2e-4 (learning rate) with a 5e-3 weight decay, CLAM-sb (CLAM) at 2e-4 (learning rate) with a 1e-3 weight decay, and DTFD at 1e-4 (learning rate) with a 1e-4 weight decay. These parameters were broadly consistent with those defined in the original implementations of these methods. All experiments are performed on RTX 3080ti 12GB GPU.

\section{Results}

\begin{figure}[h]
    \centering
    \begin{subfigure}{0.45\textwidth}
        \centering
        \includegraphics[width=\linewidth]{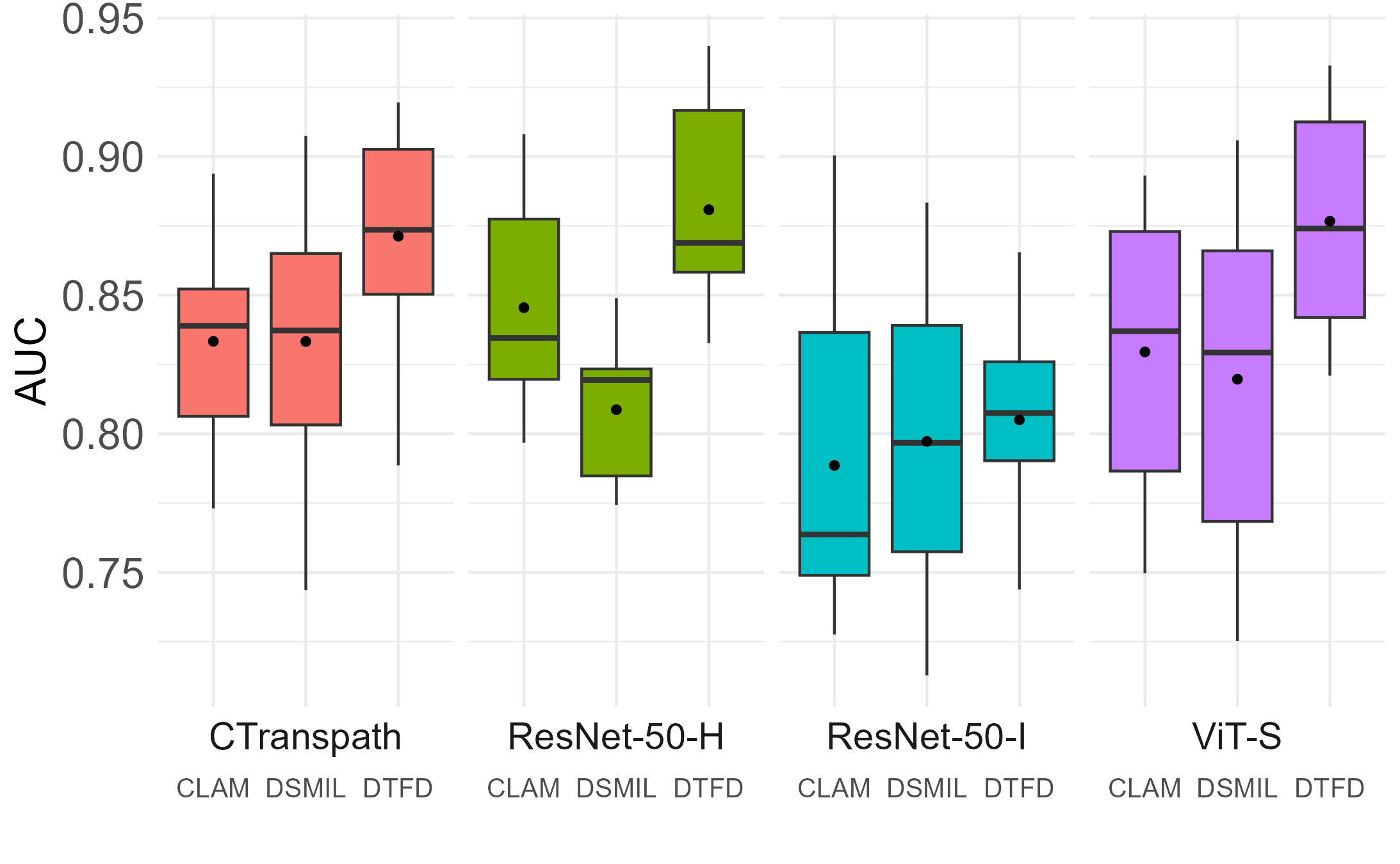}
        \caption{Area Under Curve (AUC) Scores}
    \end{subfigure}
    \hfill
    \begin{subfigure}{0.45\textwidth}
        \centering
        \includegraphics[width=\linewidth]{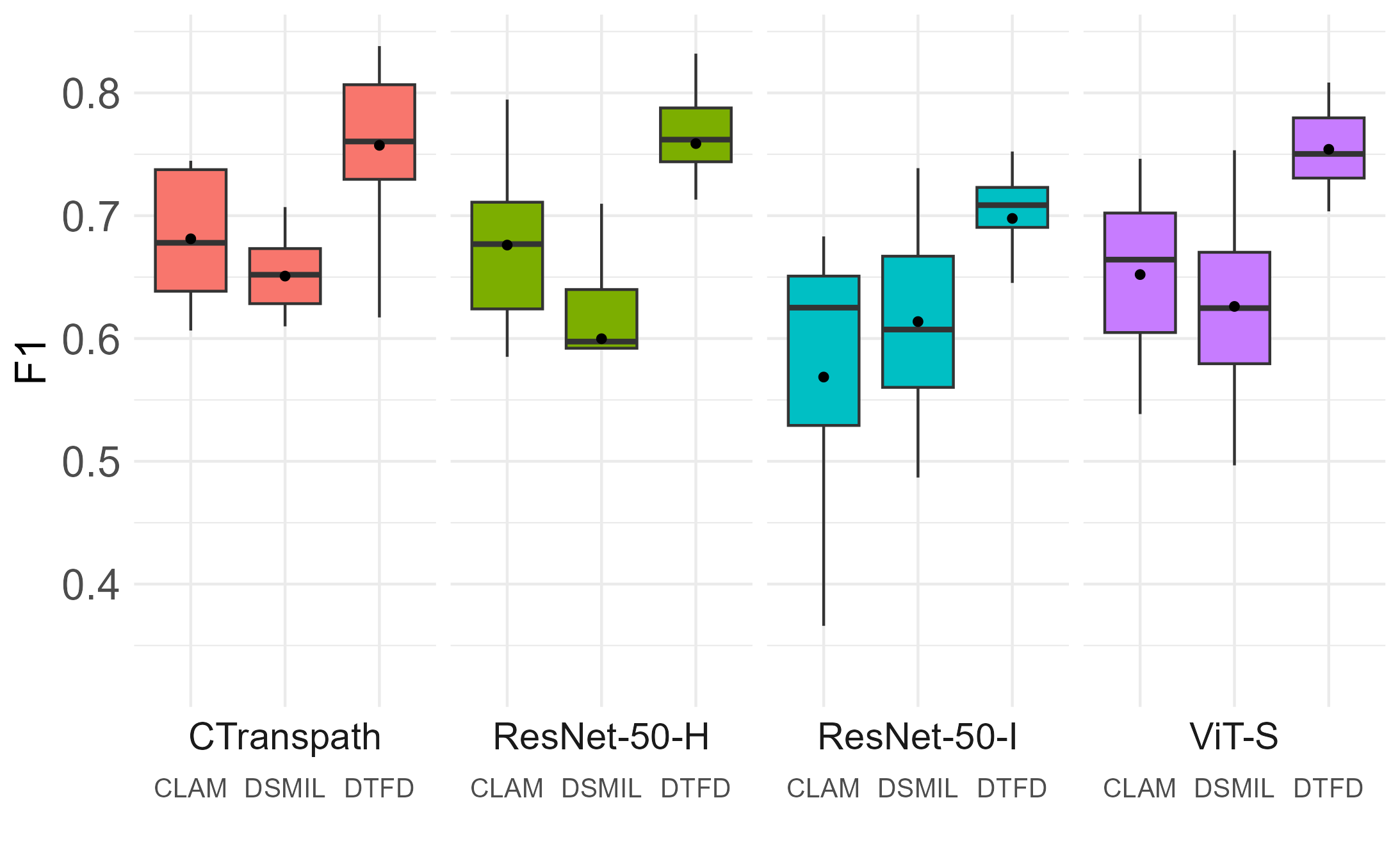}
        \caption{Macro F1 Scores}
    \end{subfigure}
    \caption{Comparative Analysis of Feature Extractors and Aggregation Methods. The box plots illustrate the 10-fold AUC and macro F1 scores distribution for all feature extractor and aggregation combinations used. Mean values are indicated by dots within each box, whereas lines represent medians.}
    \label{fig:10_fold}
\end{figure}

Our analysis across various datasets demonstrates the effectiveness of combining feature extractors, aggregators, and pretraining techniques. On the India Pathology Dataset Brain (IPD-Brain), Table~\ref{tab:subtype_ipd}, the combination of a Resnet-50 backbone, pretrained via the Barlow twin SSL method, and the DTFD aggregator emerged as superior, achieving an average AUC of 88.08\%, an accuracy of 79.83\%, and an F1 score of 75.87\%. Variation in AUC and F1 score across 10-folds for all the combinations can be observed in Fig.~\ref{fig:10_fold}. This configuration proved particularly adept at classifying brain tumor subtypes, showcasing its suitability for the unique challenges of histopathological data analysis. For the grade classification (LGG vs HGG) task on IPD-Brain, the same configuration achieved an AUC of 92.64\%, accuracy of 89.25\%, and F1 of 92.79\%.

% : 10-folds a) AUC (Top) b) F1 Score  (Bottom) results visualization

Table~\ref{tab:subtype_tcga} details the evaluation of The Cancer Genome Atlas (TCGA) Brain dataset, comparing CLAM (baseline), TPMIL \cite{yang2023tpmil} (current state-of-the-art), and DTFD (best on IPD-Brain). We established a new state-of-the-art, achieving a 95.81\% AUC, 86.94\% accuracy, and an 80.41\% F1 score, highlighting DTFD's superior performance on TCGA-Brain. Interestingly, our reimplementation of CLAM using default parameters surpassed TPMIL's documented results but not DTFD, emphasizing DTFD's adaptability and performance generalization across various histopathological datasets.

\begin{table}[h]
\centering
\caption{Tumor subtype classification results on TCGA-Brain}
\label{tab:subtype_tcga}
\begin{tabular}{|c|l|l|l|}
\hline
Feature   Aggregator & \multicolumn{1}{c|}{AUC} & \multicolumn{1}{c|}{ACC} & \multicolumn{1}{c|}{F1} \\ \hline
CLAM & 95.62 $\pm$ 2.04 & 85.62 $\pm$ 4.25 & 79.69 $\pm$ 5.86 \\ \hline
$\text{TPMIL}^*$ & 94.17 & 83.16 & -- \\ \hline
\textbf{DTFD} & \textbf{95.81 $\pm$ 1.78} & \textbf{86.94 $\pm$ 3.90} & \textbf{80.41 $\pm$ 4.38} \\ \hline
\end{tabular}
\begin{tablenotes}[]
\item   \begin{small}\textbf{$^*$}: Results are taken from the original paper.
\end{small} 
\end{tablenotes}
\end{table}

Interestingly, applying stain normalization using StainNet \cite{stainnet} on both IPD-Brain and TCGA-Brain datasets, which can be observed in Table~\ref{tab:stainNorm}, did not improve performance. It resulted in a decrease in AUC and F1 scores for both datasets. This outcome suggests that the employed feature extractor was already capable of effectively handling stain variability inherent in histopathological images. Consequently, the use of stain normalization appeared to diminish model performance slightly, emphasizing the feature extraction process's intrinsic ability to accommodate stain variations without requiring additional normalization procedures.

\begin{table}[h]
\centering
\caption{Effect of Stain Normalization on both the datasets.}
\label{tab:stainNorm}
\begin{tabular}{|l|c|c|c|c|}
\hline
\multicolumn{1}{|c|}{Dataset} & \multicolumn{1}{c|}{Stain Norm} & AUC & ACC & F1 \\ \hline
\multirow{2}{*}{IPD} & $\bm{\times}$ & \textbf{88.08 $\pm$ 3.98} & \textbf{79.83 $\pm$ 3.31} & \textbf{75.87 $\pm$ 4.79} \\ \cline{2-5} 
 & $\checkmark$ & 86.34 $\pm$ 5.25 & 77.09 $\pm$ 5.47 & 70.81 $\pm$ 8.24\\ \hline
\multirow{2}{*}{TCGA} & $\bm{\times}$ & \textbf{95.81 $\pm$ 1.78} & \textbf{86.94 $\pm$ 3.90} & \textbf{80.41 $\pm$ 4.38} \\ \cline{2-5} 
 & $\checkmark$ & 93.37 $\pm$ 3.06 & 86.67 $\pm$ 4.62 & 79.96 $\pm$ 5.38\\ \hline
\end{tabular}
\end{table}

Table~\ref{tab:ihc-table} details the classification of IHC molecular biomarkers on the IPD-Brain dataset, achieving high accuracy and AUC rates for crucial biomarkers like IDH, ATRX, and TP53. This highlights the model's potential as an accessible IHC molecular biomarker identification tool directly from H\&E slides.

\begin{table}[h]
\centering
\caption{IHC biomarkers classification results on IPD-Brain}
\label{tab:ihc-table}
\begin{tabular}{|c|c|c|c|}
\hline
Task & AUC & ACC & F1 \\ \hline
IDH & 90.66 $\pm$ 5.22 & 86.65 $\pm$ 4.58 & 86 $\pm$ 4.58 \\ \hline
ATRX & 84.07 $\pm$ 6.10 & 83.61 $\pm$ 8.68 & 67.3 $\pm$ 6.49 \\ \hline
TP53 & 73.32 $\pm$ 8.63 & 77.43 $\pm$ 9.53 & 65.83 $\pm$ 10.93\\ \hline
\end{tabular}
\end{table}

Furthermore, in Table~\ref{tab:ki67-table}, the Ki-67 marker classification task showcased the model's capability to accurately predict tumor cell proliferation rates, with significant proficiency, observed at the cutoff value of 10, achieving an 89.55\% AUC and a 91.17\% F1 score. This proficiency in Ki-67 classification underlines the model's utility in providing vital prognostic information necessary for effective treatment planning and patient management.

\begin{table}[h]
\centering
\caption{Ki-67 as classification task results}
\label{tab:ki67-table}
\begin{tabular}{|c|c|c|c|c|}
\hline
Task & Cutoff Value & AUC & ACC & F1 \\ \hline
\multirow{3}{*}{Ki-67} & 5 & 86.76 $\pm$ 6.92 & 91.73 $\pm$ 6.56 & 92.23 $\pm$ 2.19\\ \cline{2-5} 
 & 10 & 89.55 $\pm$ 5.27 & 86.93 $\pm$ 4.79 & 91.17 $\pm$ 3.33 \\ \cline{2-5} 
 & 20 & 86.81 $\pm$ 6.82 & 83.4 $\pm$ 6.65 & 85.99 $\pm$ 6.38 \\ \hline
\end{tabular}
\end{table}

Our findings demonstrate that a ResNet-50 model pre-trained with Barlow Twin SSL and combined with DTFD aggregation enhances morphological and immunohistochemical analysis in histopathology, especially in brain tumor classification. This approach underscores the effectiveness of careful feature extraction method selection, showing that proper pretraining and aggregation negate the need for stain normalization and improves diagnostic accuracy.

\section{Explainability}

\begin{figure}[h]
    \centering
    \includegraphics[width=0.45\textwidth]{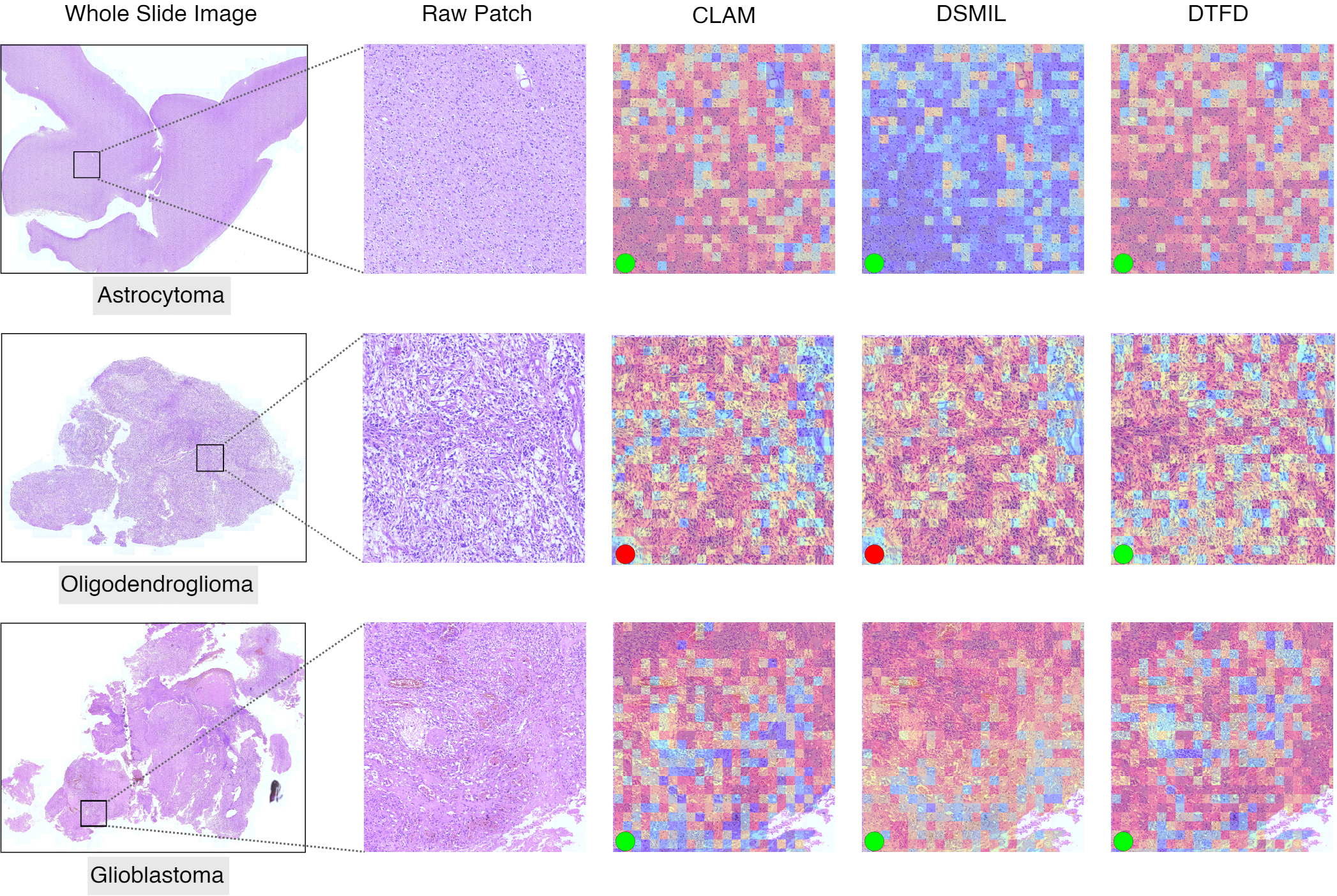}
    \caption{Attention heatmap visualization for brain tumor WSI classification, with green and red dots indicating correct classifications and misclassifications, respectively, across all feature aggregators.}
    \label{fig:interpretability}
\end{figure}

Fig.~\ref{fig:interpretability} enhances the interpretability of feature aggregators across glioma subtypes by analyzing attention weights assigned to tissue patches. This visualization highlights critical tissue patches that guide the glioma subtype classification. By employing a gradient scale from blue to red, our model signifies the importance of specific regions within the WSI, with blue indicating regions of lesser importance and red highlighting the most critical areas. In comparison, the traditional manual microscopic examination by neuropathologists in clinical settings is an exhaustive process, requiring extensive scrutiny of the entire slide to identify significant glioma morphological patterns.

Our model's practical applicability and accuracy were further substantiated through a rigorous single-blind evaluation conducted with an experienced neuropathologist from the Department of Pathology at Nizam’s Institute of Medical Sciences, a premier tertiary hospital in India. The pathologist was presented with a WSI as our model. The pathologist scrutinized specific regions within the WSIs to formulate their diagnoses, some of which are shown as 'Raw Patch' in Fig.~\ref{fig:interpretability}.

By qualitatively analyzing, we can observe DTFD's superior ability to focus on regions indicative of each glioma's medical characteristics, aligning with known pathological features. The key diagnostic features of glioblastoma, such as microvascular proliferation and palisading necrosis, were accurately highlighted by DTFD. In astrocytoma, the emphasis is on cellular structures absent of necrosis, showcasing gemistocytic features. The classic histopathologic features of oligodendroglioma, like the "chicken wire" vascular pattern and "fried egg" appearance of the cells, are also accurately identified. This assessment underlines DTFD's effectiveness in pinpointing diagnostically relevant regions, enhancing model transparency and reliability in histopathological evaluations.

The model demonstrated proficiency in identifying critical morphological microstructures, showing a strong alignment between its focus on important regions and the key areas within the raw WSIs that are essential for glioma subtype classification. This congruence closely reflects the diagnostic strategies employed by pathologists, underscoring the model's ability to prioritize and analyze patch segments in a manner similar to how experts identify glioma patterns.

\section{Conclusion}
Our study demonstrated how MIL-based algorithms can significantly improve brain tumor diagnosis by classifying subtypes, grading, and IHC biomarkers. We achieved significant improvements in diagnostic precision across various histopathological datasets employing a ResNet-50 architecture pre-trained with the Barlow Twin SSL approach and utilizing DTFD for feature aggregation. Experimentation with diverse configurations of feature extraction and aggregation led to this setup that outperformed current state-of-the-art solutions. Our methodology focuses on using H\&E stained slides exclusively, aiming to improve diagnostics in settings where IHC staining may be logistically or financially challenging. We plan to expand our research to include a wider range of feature extractors and aggregators, exploring or developing optimal combinations that could further improve diagnostic performance.

\section{Compliance with Ethical Standards}
Procedures in studies with human participants adhered to ethical standards set by institutional (NIMS-IEC) and/or national research committees (ICMR).

\section*{References}

\def\refname{\vadjust{\vspace*{-2.5em}}} %Please don't do this in a real paper.
\bibliographystyle{unsrt}

\bibliography{ref}

\end{document}